%% file: main.tex
\documentclass[letterpaper, 10 pt, conference]{ieeeconf}  

\IEEEoverridecommandlockouts                              

\usepackage{textcomp}
\usepackage[utf8]{inputenc} 
\usepackage[T1]{fontenc}    
\usepackage{hyperref}       
\usepackage{url}            
\usepackage{booktabs}       
\usepackage{amsfonts}       
\usepackage{nicefrac}       
\usepackage{microtype}      
\usepackage{xcolor}         
\usepackage{multirow}
\usepackage{amsmath}
\usepackage{amssymb}
\usepackage{makecell}
\usepackage{bm}
\usepackage[pdftex]{graphicx}
\usepackage{bm}
\usepackage{gensymb}
\usepackage[table]{xcolor} 

\title{\bf Leveraging Visual and Geometric Priors for Metric-scale and Complete Vehicle Gaussian Reconstruction from Limited Views}

\author{Jinyu Miao$^{1,\dagger}$, Jiusi Li$^{1,\dagger}$, Yifei He$^{1}$, Miao Long$^{1}$, Kun Jiang$^{1,*}$, Mengmeng Yang$^{1,*}$, Diange Yang$^{1,*}$
\thanks{This work was supported in part by the National Natural Science Foundation of China (52394264, 52472449, U22A20104, 52372414, 52402499), and Independent Research Project of the State Key Laboratory of Intelligent Green Vehicle and Mobility, Tsinghua University (ZZ-PY-20250408).}
\thanks{$^{1}$Jinyu Miao, Jiusi Li, Yifei He, Miao Long, Kun Jiang, Mengmeng Yang, and Diange Yang are with the School of Vehicle and Mobility and the State Key Laboratory of Intelligent Green Vehicle and Mobility, Tsinghua University, Beijing, China. {\tt\small jinyu.miao97@gmail.com}}%
\thanks{$^{\dagger}$The authors contribute equally to this work.}
\thanks{$^{*}$Corresponding author: Diange Yang, Kun Jiang, and Mengmeng Yang}%
}

\begin{document}

\maketitle


\input{contents/0-abstract}
\input{contents/1-intro}

\input{contents/2-related}
\input{contents/3-method}
\input{contents/4-experiment}

\input{contents/5-conclusion}

\bibliographystyle{ieeetr}
\bibliography{ref}

\end{document}

%% file: contents/0-abstract.tex
\begin{abstract}
High-fidelity vehicle assets are essential for controllable traffic scene generation, particularly for synthesizing rare and safety-critical long-tail scenarios. However, reconstructing a reusable vehicle representation from in-the-wild onboard images remains challenging for two reasons. First, image-to-3D generation methods generally produce models without reliable metric scale. Second, onboard cameras usually observe only one side of a target vehicle, making conventional multi-view reconstruction incomplete on unobserved regions. 
To solve these problems, we propose a feed-forward vehicle asset reconstruction method, which leverages two complementary priors to reconstruct 3D Gaussian representations for vehicles using sparse one-sided observations. 
To achieve metric-scale reconstruction, a visual foundation model is first utilized to serve as a visual prior for Gaussian initialization. The Gaussian attributes are then estimated by a learnable encoder-decoder module. A symmetry-aware cloning strategy is presented to complete the unobserved side directly in Gaussian space, which exploits the bilateral structure of vehicles as a geometric prior. Experiments on the public dataset demonstrate that the proposed method significantly outperforms existing approaches in both vehicle asset completeness and geometric accuracy.

\end{abstract}

%% file: contents/1-intro.tex
\section{Introduction}
\label{sec:intro}

The synthesis of diverse and physically plausible traffic scenes is an important tool for addressing long-tail problems in current autonomous driving area \cite{cornercases,cornercases2}. Many rare or hazardous scenes are difficult and costly to collect directly, but can be generated by editing foreground objects in previously captured scenes \cite{Subjectdrive}. This object-centric generation paradigm requires a large library of object assets with accurate metric dimensions, complete geometry, realistic appearance, and consistent rendering across viewpoints. In this work, we focus on vehicles, the most common foreground objects in autonomous driving scenarios.

\begin{figure}[!t]
    \centering
    \includegraphics[width=\linewidth]{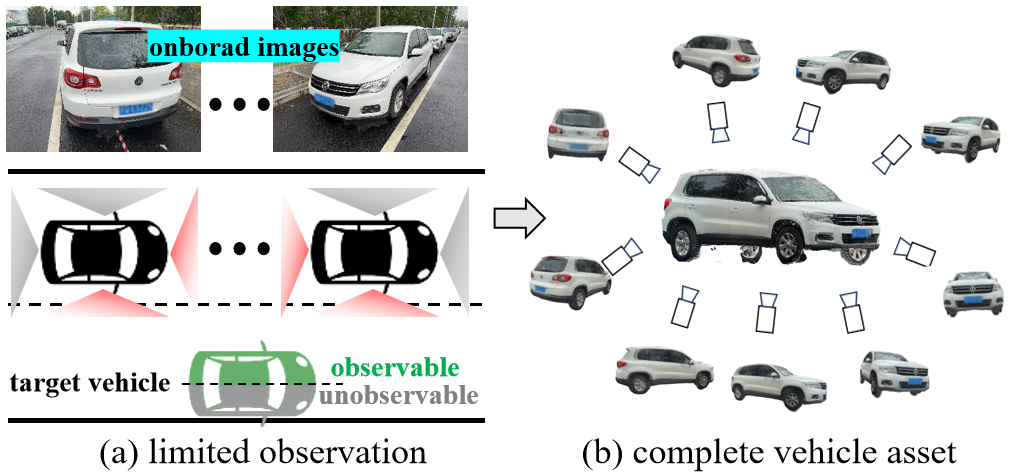}
    \caption{The objective of this work is to build a metric-scale and complete vehicle asset using limited observations from onboard images.}
    \label{fig:intro}
\end{figure}

Large-scale onboard videos contain a wide variety of vehicles, making them an attractive source for asset construction. Nevertheless, directly converting such data into reusable 3D representations is technically challenging. Existing image-to-3D generative models can hallucinate complete objects from sparse images, but their outputs usually lack reliable metric scale and may exhibit category-dependent shape or appearance bias \cite{trellis,makeit3d}. In contrast, optimization-based reconstruction methods such as 3D Gaussian Splatting (GS) can faithfully fit observed views, but they cannot recover unobserved object regions \cite{3dgs}. This limitation is especially severe for onboard imagery because an ego vehicle normally passes another vehicle along the road direction, and the target vehicle is therefore observed only within a limited range of heading angles. The opposite side of target vehicle remains largely invisible, as shown in Fig. \ref{fig:intro}.

This work addresses the problem of reconstructing a complete vehicle Gaussian representation from sparse one-sided observations. Our key idea is to combine two complementary sources of prior knowledge. First, we exploit the pretrained spatial understanding encoded in a large-scale visual foundation model. Its metric-aware multi-view estimation supplies dense metric-scale geometry, which provides stable positions and colors for initializing 3D Gaussians. Second, we exploit the bilateral symmetry of vehicles. Instead of augmenting the input images through flipping, we impose symmetry directly in the 3D Gaussian representation by cloning the well-observed side across the symmetry plane. The resulting framework uses a \emph{visual prior} to recover metric geometry and a \emph{geometric prior} to infer a complete object from sparse one-sided views.
The proposed system is feed-forward. A frozen visual foundation model reconstructs metric point clouds of target vehicle, while a trainable Gaussian head predicts Gaussian attributes. Symmetry-aware Gaussian cloning then fills the unobserved side with negligible additional computational cost. 
The main contributions are summarized as follows:
\begin{itemize}
    \item We propose a feed-forward vehicle asset reconstruction framework that combines a visual foundation model with a task-specific Gaussian head, enabling metric-scale and complete 3D Gaussian construction from onboard limited observations.
    \item We introduce a symmetry-aware Gaussian cloning strategy that applies the bilateral geometric prior directly in Gaussian space, allowing the unobserved side of target vehicle to be completed efficiently.
    \item Experiments on real data demonstrate that the proposed method achieves the best vehicle construction quality among existing baselines in terms of both geometric completeness and appearance consistency.
\end{itemize}

%% file: contents/2-related.tex
\section{Related Works}
\label{sec:related_works}

\subsection{Dynamic Traffic Scenes Reconstruction}

Owing to the novel-view rending capability, Neural Radiance Fields (NeRF) \cite{nerf} and 3D GS \cite{3dgs} have been adopted to reconstruct dynamic driving scenes and generate augmented driving data from novel viewpoints. NeRF-based methods rely on implicit scene representations and typically require auxiliary motion cues, such as optical flow \cite{suds}, temporal flow \cite{emernerf}, or scene graphs \cite{NSG}, resulting in limited computational efficiency. In contrast, 3D GS employs explicit Gaussian representations for efficient rasterization, but vanilla 3D GS assumes a static scene \cite{3dgs,sgd}. To model dynamic objects, PVG \cite{pvg} and S3Gaussian \cite{S3Gaussian} introduce temporal Gaussian attributes, while Street Gaussians \cite{streetgaussians} and DrivingGaussian \cite{drivinggaussian} separately model static backgrounds and dynamic foregrounds. These methods enable faithful scene reconstruction and realistic novel view rendering.

\subsection{Photorealistic Data Generation}

Considering the scarcity and importance of high-value long-tail data, reconstructing driving scenarios solely from onboard observations is insufficient to support the continuous improvement of autonomous driving systems. To address this limitation, some video-diffusion models guided by bird’s-eye-view (BEV) layouts are used to generate new driving data \cite{magicdrive,panacea,Subjectdrive}. However, their controllability and appearance consistency remain limited because they do not explicitly enforce geometric constraints.
More recently, scene reconstruction techniques have been extended to object-level editing for generating realistic data from real-world scenes. MARS \cite{mars} and VC-Gaussian \cite{vcgaussian}, for example, enable instance-level editing by separately modeling the foreground and background, but they often suffer from foreground–background inconsistencies caused by mismatched illumination conditions between the original scene and the edited objects. GenMM \cite{GenMM} and G\(^{2}\)Editor \cite{g2editor} alleviate this problem by employing video inpainting for object-level editing in driving videos. In these object-level editing methods, geometrically complete and appearance-consistent object assets are therefore essential for producing realistic augmented driving data.

\subsection{Object Asset Construction}

Object asset construction build complete, metric-scale object models from low-cost sensor observations. Traditional methods such as COLMAP \cite{colmap} reconstruct sparse point clouds, which are inadequate for high-quality novel-view rendering. NeRF- and 3D GS-based methods produce photorealistic object representations but require tens of minutes to reconstruct a single object \cite{dreamcar,brum,gaussianobject}. Feed-forward methods accelerate this process by predicting 3D representations in one single forward pass \cite{pixelnerf,pixelsplat,mvsplat}, yet their performance degrades with sparse onboard observations because target objects are incompletely observed. Diffusion-based methods leverage pretrained priors to complete 3D structures from one or a few views \cite{dreamfusion,zero123,trellis}, but often lack metric-scale accuracy and appearance consistency with real observations. Therefore, efficient asset construction methods capable of jointly ensuring geometric completeness, metric-scale accuracy, and appearance consistency are essential for autonomous driving data generation.

%% file: contents/3-method.tex
\section{Method}

\begin{figure*}
    \centering
    \includegraphics[width=\linewidth]{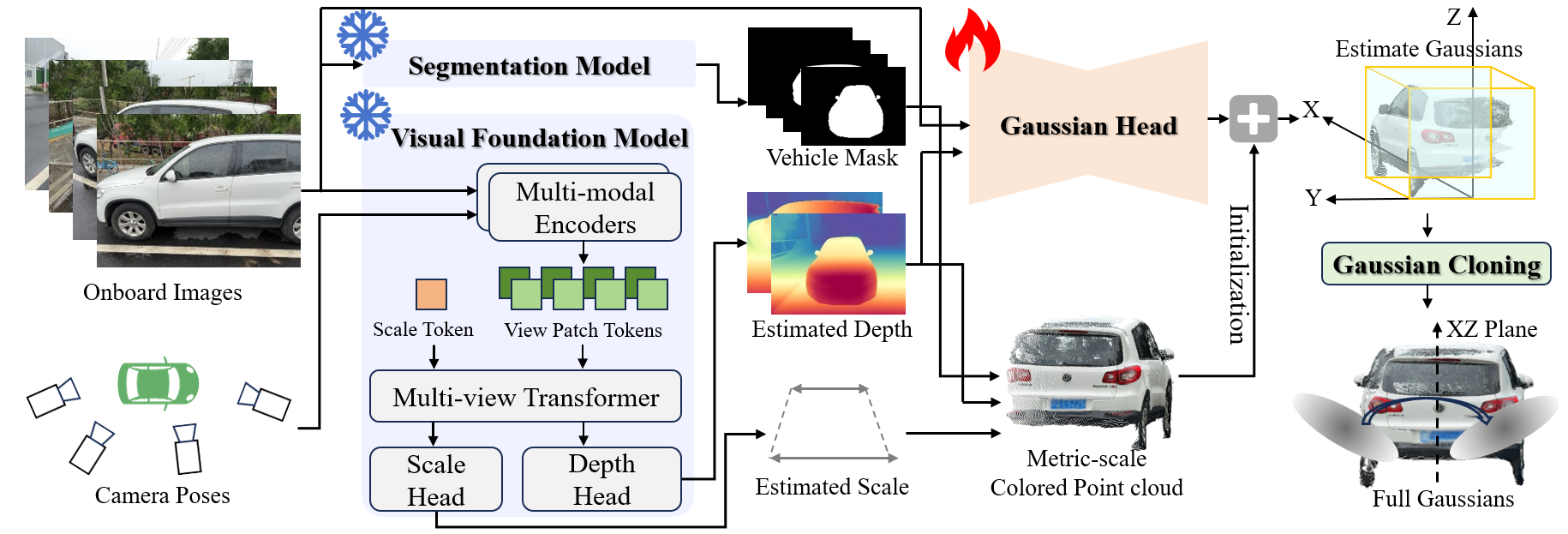}
    \caption{The overview of the proposed feed-forward vehicle Gaussians reconstruction method.}
    \label{fig:framework}
\end{figure*}

\subsection{Problem Definition and Framework Overview}

Generally, the objective of vehicle asset reconstruction task is to build a representation for a target vehicle, which has metric-scale geometry and multi-view-consistent appearance. In this work, we use 3D Gaussians as the form of asset representation due to its novel view synthesis ability. Therefore, to build the 3D Gaussian model of the target vehicle, we need to collect a set of RGB images $\mathrm{\bm{I}}= \{\bm{I}_i\}_{i=1}^{N}$ and their camera poses $\mathrm{\bm{T}}=\{\bm{T}^\mathrm{vc}_i\}_{i=1}^{N}$ expressed in a target vehicle frame $v$, where $N$ is the amount of collected images. Note that the utilized frame is defined as the ego frame following common autonomous driving systems, with $x$-axis forward, the $y$-axis left, and the $z$-axis upward. Given the images and corresponding poses, the proposed method aim to direct predict the 3D Gaussian representation $\mathcal{G}$ for the target vehicle in a feed-forward network $\mathcal{F}_\theta$:
\begin{equation}
    \mathcal{G} = \{g_j\}_{j=1}^{M}=\mathcal{F}_\theta(\mathrm{\bm{I}}, \mathrm{\bm{T}}),
    ~
    g_j = \left(\bm{\mu}_j,\bm{\Sigma}_j,\alpha_j,\bm{c}_j\right),
\end{equation}
where $\bm{\mu}_j\in\mathbb{R}^3$ is the position of Gaussian center, $\bm{\Sigma}_j\in\mathbb{R}^{3\times 3}$ is the covariance matrix, $\alpha_j\in[0,1]$ is the opacity, and $\bm{c}_j$ denotes spherical-harmonic (SH) appearance coefficients. The covariance $\bm{\Sigma}_j$ is parameterized by a rotation matrix $\bm{R}_j$ and a diagonal scale matrix $\bm{S}_j$: $\bm{\Sigma}_j = \bm{R}_j\bm{S}_j\bm{S}_j^{\top}\bm{R}_j^{\top}.$
With these estimated Gaussian attributes, we can efficiently render the appearance of the target vehicle at any viewpoints using $\alpha$-rendering techniques \cite{3dgs}. 

Considering the utility of vehicle assets, the reconstructed representation of target vehicle should both have metric-scale geometry and consistent appearance from arbitrary viewpoints so that it can be inserted into any driving scenarios to derive simulated data. 
To this end, we leverage a visual fundamental model to recover the metric-scale geometry of the target vehicle from multi-view observations and then estimate its Gaussian attributes by a learnable network as shown in Fig. \ref{fig:framework}.
Additionally, the visual observations captured by onboard camera is generally sparse and only cover one side of the target vehicle, as shown in Fig. \ref{fig:intro}. The limited viewpoints of observations challenge the complete reconstruction of target vehicle.
Therefore, we design a heuristic Gaussian cloning strategy based on the symmetry characteristic of vehicles to complete the unobserved region of Gaussian model.  

\subsection{Metric-Scale Geometry Reconstruction using Visual Prior}
\label{sec:visual}

The first step to build a vehicle asset is to recover its metric-scale geometry. Traditional rule-based multi-view stereo (MVS) algorithms \cite{colmap} fails to convergence when visual observations are too sparse to provide sufficient and accurate local feature correspondences across images. Direct predicting geometry with a learnable network offers a more generalizable approach, but the recovery of absolute scale and the requirements of large-scale training data pose extra challenges in real-world applications. Therefore, we instead leverage a frozen visual foundation model, MapAnything \cite{mapanything}, to estimate metric-scale depth from multi-view images so that the visual prior from large-scale model training can be incorporated into the vehicle asset task.  

Specifically, given multi-view images $\mathcal{I}$ and optional camera information (\textit{i.e.}, camera poses $\mathcal{T}$ in this work), the MapAnything model \cite{mapanything} predicts a dense ray map $\bm{r}_i(u,v)\in\mathbb{R}^3$ and a per-ray depth map $d_i(u,v)\in\mathbb{R}_+$ for each image, and a globally consistent scale factor $s \in \mathbb{R}_+$:
\begin{equation}
    \{d_i,r_i\}^{N}_{i=1},s=\texttt{MapAnything}(\mathrm{\bm{I}}, \mathrm{\bm{T}})
\end{equation}
where $u,v$ is the position of a pixel in the image plane. Then, a metric point $p^\mathrm{c}$ in the $i$-th camera frame $c$ is
\begin{equation}
    \bm{p}^\mathrm{c}_{i}(u,v)
    = s\,d_i(u,v)\,\bm{r}_i(u,v).
    \label{eq:metric_point_ray}
\end{equation}
Using the corresponding camera pose $T_i=(\bm{R}_{i}^\mathrm{vc},\bm{t}_{i}^\mathrm{vc})\in\mathbb{SE}(3)$ ( \textit{i.e.}, the pose transformation from the $i$-th camera frame $c$ to the vehicle frame $v$), we can obtain a metric 3D point in the vehicle frame:
\begin{equation}
    \bm{p}^\mathrm{v}_{i}(u,v)
    = \bm{R}_{i}^\mathrm{vc}\bm{p}^\mathrm{c}_{i}(u,v)
    + \bm{t}_{i}^\mathrm{vc}.
    \label{eq:vehicle_point}
\end{equation}
A vehicle segmentation mask $M_i(u,v)$ is used to remove background pixels. Then, we aggregating valid points from all input views, and yields a metric colored point cloud of the target vehicle:
\begin{equation}
    \mathcal{P}
    = \bigcup_{i=1}^{N}
    \left\{\left(\bm{p}^\mathrm{v}_{i}(u,v), \bm{I}_i(u,v)\right)
    \mid M_i(u,v)=1\right\}.
\end{equation}
This resulting colored point cloud $\mathcal{P}$ provides strong anchors for the initialization of Gaussian centers and colors. 
By using the visual foundation model pretrained on large-scale data, the training burden of the vehicle-specific network can be reduced,  and it allows the reconstructed asset to be inserted into a metric scene without post-hoc scale recovery.

\subsection{Gaussian Attribute Estimation}

The visual foundation model provides reliable geometry, but a renderable 3D Gaussian representation additionally requires opacity, anisotropic scale, orientation, view-dependent color, and local geometric refinement. Therefore, we introduce a trainable Gaussian head with a ResNet-50 \cite{resnet} encoder and a UNet-style \cite{unet} decoder. The input RGB image and estimated depth map are concatenated along the channel dimension. 
For each pixel within the vehicle, the head predicts a Gaussian attribute vector $\bm{a}_i(u,v) = \left[o,\Delta\bm{p},\bm{q},\bm{s},\bm{h}\right]$,
where $o\in\mathbb{R}$ is the raw opacity, $\Delta\bm{p}\in\mathbb{R}^3$ is a position offset of the Gaussian center, $\bm{q}\in\mathbb{R}^4$ is a rotation quaternion, $\bm{s}\in\mathbb{R}^3$ is the raw scale, and $\bm{h}\in\mathbb{R}^{27}$ contains SH coefficients up to degree 2. 
Then, the final Gaussian opacity is normalized by $\alpha = \texttt{sigmoid}(o+3)$.
The final Gaussian scale is constrained to be positive and numerically bounded:
\begin{equation}
    \bm{\sigma}
    = \texttt{clip}\left(\texttt{softplus}(\bm{s}),\sigma_{\min},\sigma_{\max}\right),
    \bm{S}=\texttt{diag}(\bm{\sigma}).
    \label{eq:scale}
\end{equation}
where $\texttt{clip}()$ truncates the out-of-bound value and $\texttt{diag}{}$ builds a diagonal matrix. The quaternion $\bm{q}$ is normalized and converted into the rotation matrix $\bm{R}^{v}$.

Rather than replacing the metric point estimated by the visual fundamental model, the head predicts offsets $\Delta\bm{p}$ of the recovered geometry so that the positions of the Gaussian centers can be corrected as:
\begin{equation}
\label{equ:offset}
    \bm{\mu}
    = \bm{p}^{v}_{i}(u,v) + \bm{R}_{i}^\mathrm{vc}\Delta\bm{p},
\end{equation}
This residual formulation preserves the global metric geometry provided by the visual foundation model while allowing the decoder to correct local depth errors around fine structures such as wheels, lamps, and mirrors of the vehicle.

As described in Sec. \ref{sec:visual}, the colors of Gaussians are initialized using the observed RGB color. Therefore, the estimated 0-order SH coefficient is adjusted by
\begin{equation}
    \bm{h}_{0}
    \leftarrow \bm{h}_{0} + \operatorname{RGB2SH}(\bm{I}_i),
\end{equation}
where
\begin{equation}
    \operatorname{RGB2SH}(\bm{x})
    = 2\sqrt{\pi}(\bm{x}-0.5).
\end{equation}
This initialization allows the representation to reproduce the basic vehicle color before learning higher-order view-dependent effects, which facilitates the network training.

\subsection{Symmetry-Aware Gaussian Cloning using Geometry Prior}

Sparse onboard observations generally cover only one side of a target vehicle, as shown in Fig. \ref{fig:intro}. A purely data-driven representation therefore estimates dense Gaussians on the observed side but only sparse or unreliable geometry on the unobserved side. To address this issue, we exploit the approximate bilateral symmetry of the common vehicles as an explicit geometric prior.

As shown in Fig. \ref{fig:framework}, given the defined vehicle frame, the geometry and appearance of the vehicle are generally symmetric relative to the $xz$-plane.
A native solution to complete the estimated Gaussian is to flip the observed images so that the complementary viewpoints can cover the unobserved side of the target vehicle. However, such an approach brings a significantly enlarged computational burden during both training and inference. Therefore, we instead propose to clone estimated Gaussians, which offers an effective and efficient alternative for complete vehicle asset reconstruction.

Specially, during both training and inference, we first count the Gaussians on the two sides of the $xz$-plane and select the denser side as the reference side. For every estimated Gaussian on the reference side, we create a flipped Gaussian with center $\widetilde{\bm{\mu}}$, covariance $\widetilde{\bm{\Sigma}}$, and rotation matrix $\widetilde{\bm{R}}$:
\begin{equation}
    \widetilde{\bm{\mu}}
    = \bm{M}\bm{\mu},
    \widetilde{\bm{\Sigma}}
    = \bm{M}\bm{\Sigma}
      \bm{M}^{\top},
    \widetilde{\bm{R}}=\bm{M}\bm{R}
    \label{eq:mirror}
\end{equation}
where the flip operation can be mathematically modeled as a flip matrix $\bm{M}$:
\begin{equation}
    \bm{M}
    = \begin{bmatrix}
    1 & 0 & 0\\
    0 & -1 & 0\\
    0 & 0 & 1
    \end{bmatrix}
    \label{eq:mirror_matrix}
\end{equation}
And the scale $\widetilde{\bm{S}}$ and opacity $\widetilde{{\alpha}}$ are directly cloned from the reference Gaussians. For SH coefficients, flipping across the $xz$-plane changes the sign of coefficients associated with odd azimuthal order.
Finally, the reference and flipped Gaussians are merged as a complete set of vehicle asset representation:
\begin{equation}
    \mathcal{G}_{\mathrm{full}}
    = \mathcal{G}_{\mathrm{reference}}
      \cup \widetilde{\mathcal{G}}_{\mathrm{flip}}.
\end{equation}

This Gaussian cloning strategy directly imposes the geometric prior in the output representation. It therefore avoids doubling the number of encoded views while preserving nearly the same rendering quality.

\subsection{Training scheme}

The network is trained through novel-view rendering supervision. During each iteration, $N$ images from one side of a vehicle are used as input. Two target images are randomly sampled from the opposite side. To avoid trivial correspondence with horizontally mirrored inputs, target cameras whose positions are within 0.5 meters of a mirrored input pose are excluded.

The predicted full Gaussian representation is rendered at the viewpoint of each target camera to obtain $\widehat{\bm{I}}_k$. The training objective combines pixel-wise mean squared error (MSE) and learned perceptual image patch similarity (LPIPS) \cite{Zhang2018TheUE}:
\begin{equation}
    \mathcal{L}
    = \sum_{k=1}^{}
    \left[
    \operatorname{MSE}(\widehat{\bm{I}}_k,\bm{I}_k)
    + 0.05 \cdot
    \operatorname{LPIPS}(\widehat{\bm{I}}_k,\bm{I}_k)
    \right],
    \label{eq:loss}
\end{equation}

Inspired by Flash3D \cite{flash3d}, the visual foundation model is frozen throughout training, and only the Gaussian attribute estimation network are optimized. This partial-freezing strategy preserves the visual prior information of pretrained model while reducing trainable parameters, memory consumption, and training time.

%% file: contents/4-experiment.tex
\section{Experiments}
\subsection{Dataset and Preprocessing}
We evaluate the proposed method on the 3DRealCar dataset \cite{3drealcar}.
The preprocessing pipeline follows the standard dataset protocol. COLMAP \cite{colmap} is first used to reconstruct a point cloud and camera poses up to an unknown scale. After transform the estimated camera pose to the ego frame of target vehicle, we align the COLMAP reconstruction to the scanned point cloud so as to scale the camera poses to metric units.
A total of 45 vehicles with diverse shapes and appearances are used in this work. 25 vehicles form the training set and the other 20 vehicles form the test set. The resolution of rendered images are set to $512\times384$.

\begin{figure*}[!t]
    \centering
    \includegraphics[width=\linewidth]{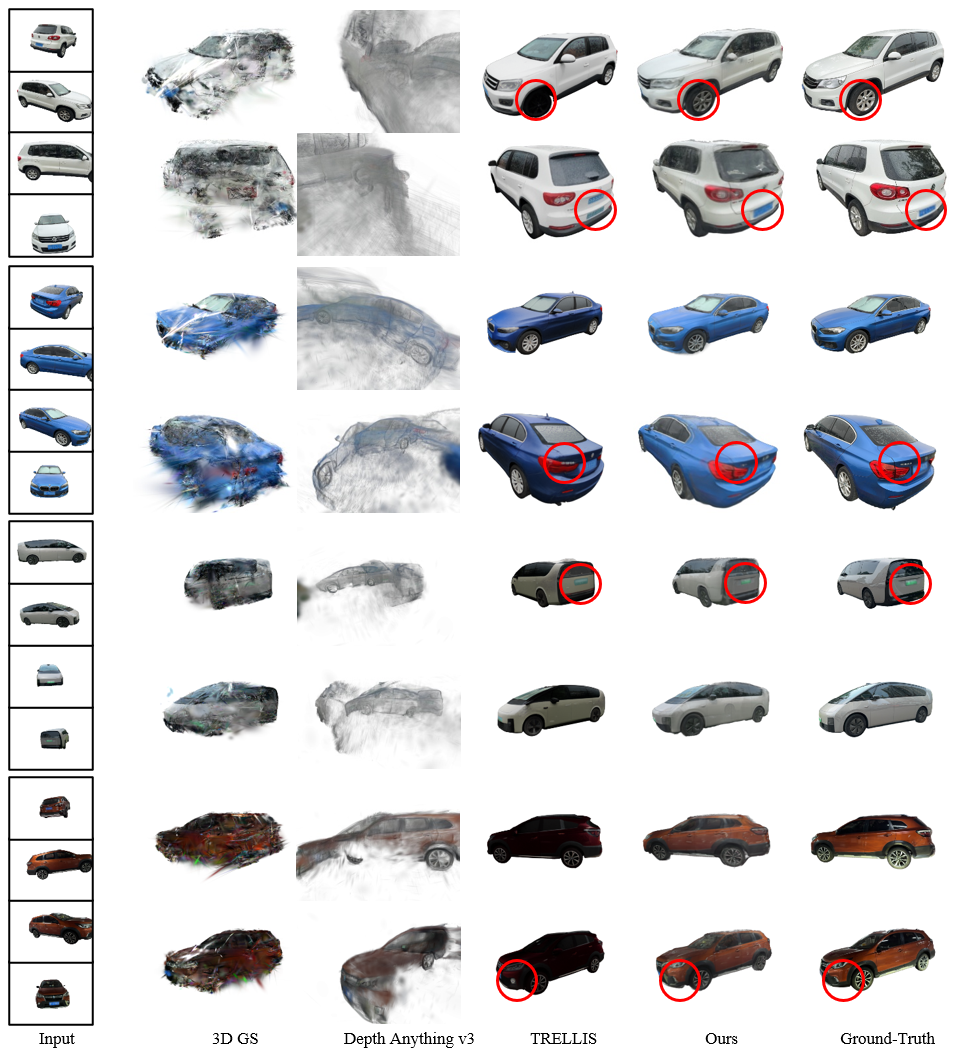}
    \caption{Qualitative comparison with existing methods in terms of opposite view rendering.}
    \label{fig:comparison}
\end{figure*}

\subsection{Experimental Setup}
For each test vehicle, 4 images from one side are used to construct the 3D Gaussian model, which simulates the limited views inherent in real-world onboard videos. 8 camera viewpoints are uniformly selected along the opposite side for novel-view rendering evaluation. We utilize peak signal-to-noise ratio (PSNR), structural similarity index measure (SSIM), and LPIPS \cite{Zhang2018TheUE} as evaluation metrics.
To prove the novelty of the proposed method, we compare against three representative baselines:
\begin{itemize}
    \item \textbf{3D GS} \cite{3dgs}, an optimization-based reconstruction method that fits the observations independently for each vehicle;
    \item \textbf{DepthAnything v3 (DA3)} \cite{da3}, a feed-forward multi-view depth and Gaussian reconstruction method. We utilize the public available DA3NESTED-GIANT-LARGE model. The estimated per-image Gaussians are filtered by semantic segmentation model \cite{sam} and then aggregated;
    \item \textbf{TRELLIS} \cite{trellis}, a generative image-to-3D model conditioned on multiple input images. Since TRELLIS \cite{trellis} does not directly produce metric-scale outputs, its predicted Gaussians are aligned to a metric COLMAP point cloud through a scale factor before evaluation.
\end{itemize}

The proposed model is trained on a NVIDIA Tesla V100 GPU with a batch size of 4 vehicles for 300 iterations. AdamW optimizer is used with a learning rate of $10^{-4}$. The visual foundation model is initialized by MapAnything \cite{mapanything} pretrained weights and keeps frozen during training. The scale bounds in Eq.~\eqref{eq:scale} are set to $\sigma_{\min}=10^{-10}$ and $\sigma_{\max}=3$.

\subsection{Comparison with State-of-the-Art Methods}

Figure \ref{fig:comparison} and Table~\ref{tab:main_results} compare the opposite-side novel-view rendering performance of the evaluated methods under the same sparse, one-sided input. Optimization-based 3D GS \cite{3dgs} produces substantial geometric incompletion and blurred textures, indicating that direct optimization mainly interpolates observed content and struggles to infer unobserved structures. DA3 \cite{da3} also fails to recover complete fine-grained geometry because of its limited Gaussian decoder and inaccurate pose estimation, which lead to misaligned depth across views and distorted 3D Gaussian models. TRELLIS \cite{trellis} generates more complete shapes and achieves better SSIM and LPIPS than reconstruction-based methods, but its category-level generation prior does not preserve the metric scale and instance-specific appearance of the observed vehicle. Consequently, its generated assets lack instance-specific appearance details and exhibit limited visual fidelity to the real vehicle.
In contrast, our method achieves the best performance on all three metrics, with a PSNR of 19.030 dB, an SSIM of 0.789, and an LPIPS of 0.183, corresponding to a 21.65\% improvement in PSNR, a 10.35\% improvement in SSIM, and an 18.67\% reduction in LPIPS relative to the strongest competing results. It reconstructs geometrically complete opposite-side structures with rich surface details and appearance consistent with the observations. Moreover, its feed-forward inference directly produces a complete, metric-scale 3D Gaussian vehicle model from given views, avoiding time-consuming per-instance optimization and post-processing scale adjustment, thereby providing an efficient and directly usable asset for object-controllable scene generation.

\begin{table}[!t]
\centering
\caption{Opposite-side novel-view rendering on 3DRealCar.}
\label{tab:main_results}
\resizebox{\linewidth}{!}{
    \begin{tabular}{lccc}
    \toprule
    Method & PSNR $\uparrow$ & SSIM $\uparrow$ & LPIPS $\downarrow$ \\
    \midrule
    3D GS \textcolor{gray}{ToG 2023} \cite{3dgs} & 15.643 & 0.675 & 0.294 \\
    DA3 \textcolor{gray}{ICLR 2026} \cite{da3} & 10.625 & 0.589 & 0.500 \\
    TRELLIS \textcolor{gray}{CVPR 2025} \cite{trellis} & 14.541 & 0.715 & 0.225 \\
    \rowcolor{gray!20}Ours & \textbf{19.030} & \textbf{0.789} & \textbf{0.183} \\
    \bottomrule
    \end{tabular}
}
\end{table}

We additionally evaluate the robustness of the proposed method to the input observation side. As shown in Table~\ref{tab:side_robustness}, left- and right-side input views yield comparable opposite-side rendering quality, indicating that the method is not tied to a specific observation direction. It can effectively exploit the inherent symmetry prior of vehicles to complete the geometry and appearance of the unobserved side from a single-side observation. The full-surround setting achieves slightly better scores because it includes both observed and unobserved regions, whereas opposite-side evaluation relies more heavily on geometric completion.

\begin{table}[!t]
\centering
\caption{Robustness to the side of the input observations.}
\label{tab:side_robustness}
\resizebox{\linewidth}{!}{
\begin{tabular}{llccc}
\toprule
Input View & Test View & PSNR $\uparrow$ & SSIM $\uparrow$ & LPIPS $\downarrow$ \\
\midrule
Right & Left & 19.030 & 0.789 & 0.183 \\
Left & Right & 18.814 & 0.792 & 0.173 \\
Right & Left+Right & 19.719 & 0.816 & 0.158 \\
Left & Left+Right & 19.607 & 0.818 & 0.159 \\
\bottomrule
\end{tabular}
}
\end{table}

\subsection{Ablation Studies}
\subsubsection{Gaussian Position Offset}
In this work, we initialize the Gaussian centers using geometry estimates from a visual foundation model and subsequently refine them with the position offsets predicted by the Gaussian head, as defined in Eq.~\eqref{equ:offset}. To evaluate the effectiveness of this position-refinement scheme, we compare the proposed method with and without the predicted offsets. The results are reported in Table \ref{tab:offset_ablation}. Directly using the initial geometry estimated by the visual foundation model already achieves satisfactory performance without position refinement, demonstrating the value of the visual prior. However, geometry recovered from multi-view depth estimates inevitably contains mismatches and cannot perfectly capture the fine-grained geometry of the target vehicle. After introducing the learnable offsets, all evaluation metrics consistently improve. These results indicate that the initial Gaussian positions constrain the representation of detailed geometry, whereas learnable position refinement enables the Gaussians to better capture fine-grained geometric structures, thereby improving novel-view rendering quality.

\begin{table}[!t]
\centering
\caption{Effectiveness of Gaussian position offset estimation.}
\label{tab:offset_ablation}
\resizebox{\linewidth}{!}{
\begin{tabular}{lccc}
\toprule
Setting & PSNR $\uparrow$ & SSIM $\uparrow$ & LPIPS $\downarrow$ \\
\midrule
\textit{w.o.} position offset & 18.342 & 0.777 & 0.192 \\
\textit{w.} position offset & \textbf{19.030} & \textbf{0.789} & \textbf{0.183} \\
\bottomrule
\end{tabular}
}
\end{table}

\begin{table}[!t]
\centering
\caption{Comparison of two Gaussian completion methods using symmetry-aware geometric prior.}
\label{tab:symmetry_ablation}
\resizebox{\linewidth}{!}{
\begin{tabular}{lcccc}
\toprule
Method & PSNR $\uparrow$ & SSIM $\uparrow$ & LPIPS $\downarrow$ & Time (s) $\downarrow$ \\
\midrule
Image flipping \cite{dreamcar} & \textbf{19.133} & \textbf{0.792} & \textbf{0.182} & 3.032 \\
\rowcolor{gray!20}Gaussian cloning & 19.030 & 0.789 & 0.183 & \textbf{1.547} \\
\bottomrule
\end{tabular}
}
\end{table}

\subsubsection{Gaussian Cloning \textit{v.s.} Image Flipping}

To complete the geometry and appearance of unobserved regions from limited single-sided views, we exploit the symmetry-aware geometric prior of vehicles to clone Gaussians. To evaluate this strategy, we compare it with the commonly used image-level flipping approach \cite{dreamcar}. This approach horizontally flips the observed images to simulate views from the opposite side and feeds both the original and flipped images into the reconstruction network. The results are reported in Table \ref{tab:symmetry_ablation}. Image flipping achieves slightly higher rendering quality by providing additional image-level observation and supervision, but it doubles the number of input images and increases the reconstruction time to 3.032 s. In contrast, the proposed Gaussian-cloning strategy achieves comparable rendering quality while substantially improving efficiency, requiring only 1.547 s, which is approximately half the time required by image flipping. These results demonstrate that Gaussian cloning provides an effective balance between rendering quality and reconstruction efficiency under limited single-sided observations.

\subsubsection{Activation Functions for Scale Parameterization}

In Eq. \ref{eq:scale}, the raw scale predicted by the Gaussian head is processed to ensure positive and numerically bounded values. We compare Softplus, Sigmoid, and ReLU as nonlinear activation functions for scale parameterization of Gaussians. As shown in Table~\ref{tab:activation_ablation}, Softplus achieves the best performance. Its smooth and strictly positive output is suitable for enforcing valid Gaussian scales while facilitating stable optimization.

\begin{table}[t]
\centering
\caption{Comparison of activation functions for Gaussian scale parameterization.}
\label{tab:activation_ablation}
    \begin{tabular}{lccc}
    \toprule
    Activation & PSNR $\uparrow$ & SSIM $\uparrow$ & LPIPS $\downarrow$  \\
    \midrule
    Sigmoid & 18.966 & 0.788 & 0.188 \\
    ReLU & 18.506 & 0.744 & 0.203 \\
    \rowcolor{gray!20}Softplus & \textbf{19.030} & \textbf{0.789} & \textbf{0.183} \\
    \bottomrule
    \end{tabular}
\end{table}

\begin{figure}[!t]
    \centering
    \includegraphics[width=\linewidth]{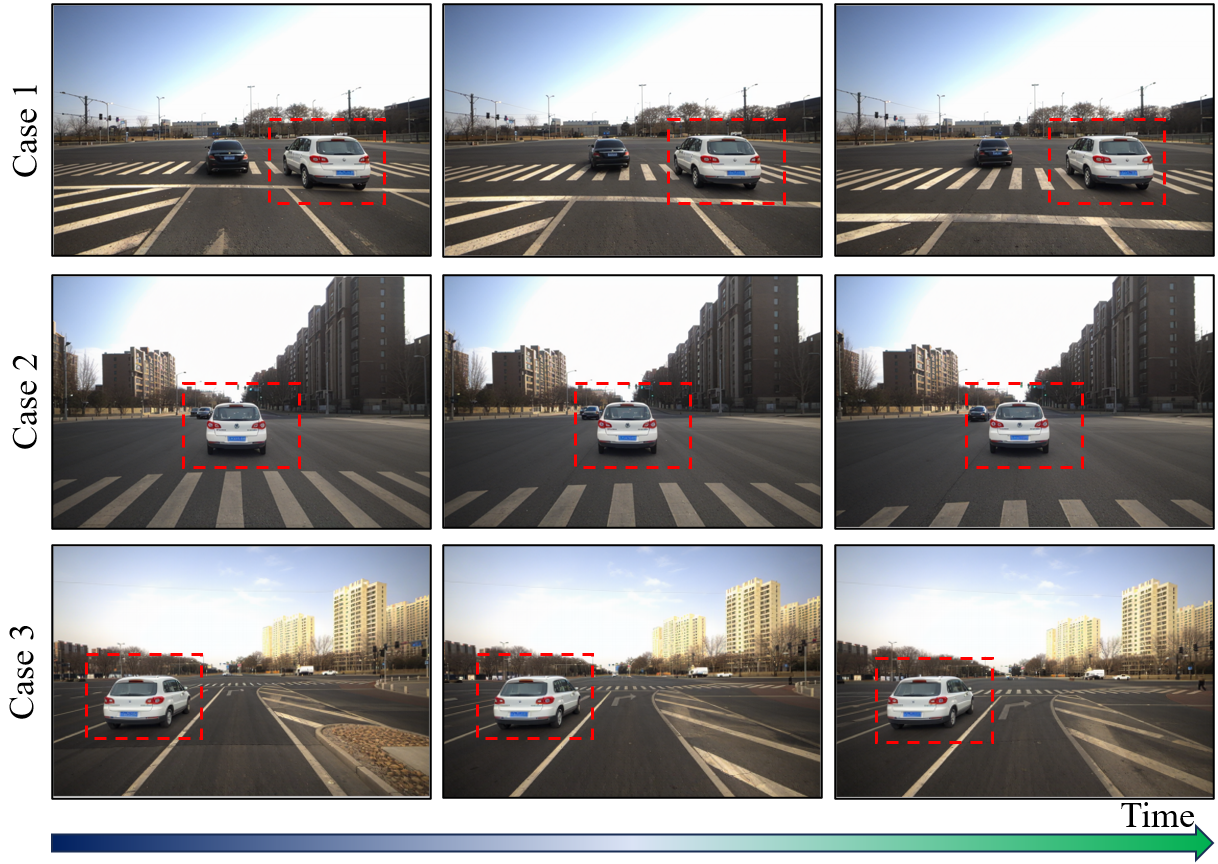}
    \caption{Object-level editing performance on real-world self-collected data.}
    \label{fig:realdata}
\end{figure}

\subsection{Real-world Object-level Editing}
The proposed method aims to provide vehicle assets that are directly pluggable, geometrically accurate, and visually realistic for object-level editing-based data generation. To validate this capability, we collect onboard videos in the Yizhuang area of Beijing, China, and insert vehicle assets reconstructed by our method into these videos using the G\(^{2}\)Editor approach \cite{g2editor}. Representative visualization results are shown in Fig. \ref{fig:realdata}. As illustrated, our method can generate geometrically complete, metric-scale, and multi-view consistent vehicle assets from limited onboard observations. These assets can therefore be seamlessly integrated with existing object-level editing methods and rendered from specified novel viewpoints to generate augmented driving data. This capability facilitates the generation of diverse long-tail and safety-critical driving scenarios, providing valuable data for improving the robustness and generalization of autonomous driving perception and planning systems.

%% file: contents/5-conclusion.tex
\section{Conclusion}
\label{sec:conclu}

In this work, we presented a feed-forward framework for reconstructing metric-scale and complete vehicle Gaussian models from limited onboard images. The method is founded upon a dual-prior framework. A large-scale visual foundation model contributes a metric-aware visual prior and supplies metric-scale Gaussian anchors, while bilateral vehicle symmetry provides a geometric prior for completing unobserved regions through direct Gaussian cloning. A trainable Gaussian head complements these two priors by estimating renderable attributes and local geometric corrections. Experiments on 3DRealCar demonstrate substantial improvements over existing baselines. The resulting assets combine metric scale, geometric completeness, multi-view consistency, and efficient construction, making them suitable for object-level editing and controllable traffic-scene generation.